\documentclass{article}

\usepackage{arxiv}

\usepackage[utf8]{inputenc} % allow utf-8 input
\usepackage[T1]{fontenc}    % use 8-bit T1 fonts
\usepackage{hyperref}       % hyperlinks
\usepackage{url}            % simple URL typesetting
\usepackage{booktabs}       % professional-quality tables
\usepackage{amsfonts}       % blackboard math symbols
\usepackage{nicefrac}       % compact symbols for 1/2, etc.
\usepackage{microtype}      % microtypography
\usepackage{lipsum}
\usepackage{graphicx}
\graphicspath{ {./images/} }

\title{Comparing AutoML and Deep Learning Methods for Condition Monitoring using Realistic Validation Scenarios}

\author{
 Payman Goodarzi \\
  Lab for Measurement Technology\\
  Saarland University\\
  Saarbrücken, Germany \\
  \texttt{p.goodarzi@lmt.uni-saarland.de} \\
   \And
 Andreas Schütze \\
  Lab for Measurement Technology\\
  Saarland University\\
  Saarbrücken, Germany \\
  \texttt{schuetze@lmt.uni-saarland.de} \\
  \And
 Tizian Schneider \\
  Lab for Measurement Technology\\
  Saarland University\\
  Saarbrücken, Germany \\
  \texttt{t.schneider@lmt.uni-saarland.de} \\
}
\usepackage{tcolorbox}
\begin{document}
\maketitle
\begin{tcolorbox}[colframe=blue!50!black,colback=blue!10!white]
\textbf{This work has been submitted to the IEEE for possible publication. Copyright may be transferred without notice, after which this version may no longer be accessible.}
\end{tcolorbox}

\begin{abstract}

This study extensively compares conventional machine learning methods and deep learning for condition monitoring tasks using an AutoML toolbox. The experiments reveal consistent high accuracy in random K-fold cross-validation scenarios across all tested models. However, when employing leave-one-group-out (LOGO) cross-validation on the same datasets, no clear winner emerges, indicating the presence of domain shift in real-world scenarios. Additionally, the study assesses the scalability and interpretability of conventional methods and neural networks. Conventional methods offer explainability with their modular structure aiding feature identification. In contrast, neural networks require specialized interpretation techniques like occlusion maps to visualize important regions in the input data. Finally, the paper highlights the significance of feature selection, particularly in condition monitoring tasks with limited class variations. Low-complexity models prove sufficient for such tasks, as only a few features from the input signal are typically needed. In summary, these findings offer crucial insights into the strengths and limitations of various approaches, providing valuable benchmarks and identifying the most suitable methods for condition monitoring applications, thereby enhancing their applicability in real-world scenarios.
\end{abstract}

% keywords can be removed

\section{Introduction}
In industrial environments, mechanical systems undergo wear and tear, which can lead to faults and breakdowns necessitating maintenance. Maintenance plays a vital role in ensuring optimal functionality and prolonging the lifespan of these systems. Reactive maintenance, the traditional approach, involves responding to faults and failures as they arise, often resulting in prolonged periods of downtime. Conversely, preventive maintenance utilizes historical data to perform regular maintenance, aiming to prevent unforeseen failures \cite{gertsbakh2000reliability}.

To overcome the limitations of these approaches, predictive maintenance has emerged as a promising alternative. By harnessing advanced machine learning techniques, predictive maintenance analyzes data from various sensors and sources to detect patterns and trends that can be utilized for predicting future failures. By identifying potential issues in advance, predictive maintenance enables timely interventions and minimizes downtime, leading to substantial cost savings and improved efficiency \cite{mobley2002introduction}.

Condition monitoring and predictive maintenance rely on the utilization of diverse sensors, such as pressure, vibration, and temperature sensors, to detect or predict forthcoming faults in industrial systems. Unlike computer vision tasks that involve extracting complex features from raw data \cite{Olah2017}, condition monitoring tasks typically rely on simpler statistical measures and a limited set of features\cite{Schneider2018}. Traditionally, effectively dealing with industrial data necessitates domain experts' expertise and a feature engineering process to extract dependable features \cite{Avci2021}.

In recent years, there has been a noticeable shift in the field of condition monitoring towards employing deep neural networks (DNN) for direct analysis of raw data, bypassing the traditional feature engineering process \cite{zhao2019}. Although this approach has shown promising results, it is not without its challenges. The inherent complexity of DNN models makes tuning their hyperparameters (HP) difficult, presenting a significant challenge. Additionally, the interpretability of DNNs decisions remains an unresolved issue \cite{Selvaraju2020}. Alternatively, the adoption of Automated machine learning (AutoML) techniques offers a viable solution. AutoML is an advanced methodology that automates various aspects of machine learning, including feature engineering, model selection, and hyperparameter optimization. Its primary goal is to enhance the accessibility of machine learning for non-experts by reducing the need for manual intervention in model development \cite{Truong2019}. AutoML algorithms can automatically explore a predefined set of models and associated HPs, or even perform neural architecture searches to discover novel model architectures that deliver optimal performance for a given task \cite{He2021}. By leveraging these algorithms, users can streamline the machine learning process and obtain high-performing models without extensive manual experimentation.

In this research, an experimental approach is employed to compare the performance of conventional AutoML methods with DNN solutions based on neural architecture search for condition monitoring and predictive maintenance. While existing studies have focused on comparing and benchmarking machine learning methods for fault detection and time series signals, our work distinguishes itself as the most comprehensive due to its utilization of real-world practical datasets. Esakimuthu Pandarakone et al.  \cite{Esakimuthu2019} examined the effectiveness of classical ML classifiers and convolutional neural networks (CNN) in detecting bearing faults in induction motors. Buckley et al. \cite{Buckley2022} conducted intriguing experiments on benchmark feature extraction and selection methods for structural health monitoring using two datasets. Fawaz et al. \cite{fawaz2019} explored various methods, including deep learning models, for time series analysis on UCR/UEA datasets. Although our work builds upon the foundation of AutoML from \cite{Schneider2018}, we have extended it by introducing new scenarios and methods.

We carefully selected a diverse set of datasets from this field, encompassing different observation scenarios with varying numbers of observations, as well as employing various validation strategies. By applying different validation strategies, our aim is to examine the impact of domain shift \cite{Goodarzi2022} on the performance of different methods. Our experimental design enables a comprehensive evaluation of the effectiveness of these approaches, providing valuable insights into the relative strengths and limitations of each method. Ultimately, our study aims to offer guidance to practitioners in selecting the most suitable method for their specific application in condition monitoring and predictive maintenance.

\section{Material and methods}
\label{methods}
\subsection{Databases and scenarios}

This study examined a range of datasets mainly in the domains of condition monitoring and predictive maintenance. The datasets were carefully selected to encompass various sizes and levels of complexity, representing both simple and large-scale use cases. Each dataset was categorized based on meaningful specifications or conditions and underwent validation using either random K-fold and leave-one-group-out (LOGO) cross-validation. We compared both validation methods. To preserve the distribution, we employed stratified resampling and ensured that the number of folds in the K-fold scenario matched the number of distinct groups in the LOGO validation scenario. We compared both validation methods to highlight the issue with domain shift in real-world scenarios. It is essential to ensure that supervised ML tasks are confined to a single domain, with training and test data originating from the same distribution. However, in practical scenarios, such as condition monitoring and predictive maintenance, the presence of various operational conditions can lead to shifts in the data distribution. These cross-influence conditions, including temperature variations, rotational speed changes, load variations, pressure fluctuations, or alterations in the working device, can significantly impact the performance and generalization of the ML models \cite{Goodarzi2022}. As a result, it becomes crucial to address these domain shift challenges to ensure the effectiveness and reliability of the predictive models in real-world industrial applications. Utilizing LOGO validation, while considering carefully selected cross-influence conditions, serves as a valuable metric to evaluate the model's generalizability across different working conditions.  Below is a concise overview of the datasets employed in this study.

The primary focus of this article is on supervised learning tasks with predetermined target values. Although some tasks may involve regression, our main emphasis revolves around the classification format of these tasks.  Additionally, to maintain simplicity and avoid complications arising from multi-modality, only a single sensor is utilized in the datasets.

\paragraph{The case western reserve university bearings (CWRU).}
The CWRU dataset \cite{CWRU} is widely recognized in the field of predictive maintenance and has been extensively utilized in numerous studies. The primary objective of this dataset is to perform a binary classification task, specifically distinguishing between various fault types and healthy devices. The classes in the dataset include "healthy," "inner ring" faults, "outer ring" faults, and "ball" faults.

The data comprises vibration signals obtained from bearings experiencing different fault types and varying load conditions. The recordings are specifically collected for different motor loads, namely 0, 1, 2, and 3 HP. To ensure consistency, the recordings in the dataset are sampled at a rate of 12 kHz, and the data is segmented into non-overlapping slices of 1k length.
\paragraph{The ZeMA hydraulic system (HS).}
HS dataset \cite{Schneider2018HS} comprises recordings from a testbed equipped with multiple sensors (17) capturing data under various fault conditions. The target variable in this dataset is the accumulator pre-charge pressure. For the purpose of LOGO validation, the cooler performance at 3\%, 20\%, and 100\% is regarded as the crucial control variable.

In this dataset, the analysis focuses solely on recordings from the first pressure sensor (PS1). Two system variables, namely the valve state and the accumulator (acm) state, are selected as the target variables. As a result, the dataset is split into two versions for this study, namely HS (valve) and HS (acm).

\paragraph{ZeMA electromechanical cylinder (EC).}
The  dataset \cite{Klein2018} consists of data collected from 11 sensors during the lifetime measurement of the cylinder. The cylinder follows a fixed working cycle of 2.8 seconds, including a forward stroke, waiting time, and a return stroke. For dataset creation, one second of the return stroke from every 100th working cycle was selected. In this dataset, the target variable is divided into five categories, ranging from '1' representing a new device to '5' representing a near-failure device. This classification task is utilized for lifetime estimation.

The dataset specifically utilizes the microphone as the input sensor, with a sampling rate of 2kHz. In the LOGO validation scenario, the performance of the models is evaluated using four different devices. This approach helps assess the generalizability and robustness of the models across various devices.
\paragraph{Open guided waves (OGW).}
The OGW dataset \cite{Moll2019} consists of time series signals that capture guided waves recorded at various temperature levels, ranging from 20°C to 60°C with 0.5°C increments. The signals were collected using twelve ultrasonic transducers arranged in a sender-receiver configuration. These transducers were attached to a Carbon Fiber Reinforced Polymer (CFRP) plate, which had a detachable aluminum mass positioned at four different locations to simulate delamination damage.

To generate the signals, a 5-cycle Hann-windowed sine wave was used as the source signal, with frequencies varying from 40kHz to 260kHz in 20kHz increments. The measurements were initially conducted on an intact CFRP plate at different temperature levels. Subsequently, the measurements were repeated with simulated damages at each of the four positions, along with measurements of the intact plate.

For the LOGO validation approach, this dataset focuses on a specific combination of transducers and includes four locations of simulated damage positions, as well as measurements of the intact plate. The objective of this use case is to detect whether the CFRP plate is damaged or intact, thereby presenting a binary classification task.

\paragraph{Paderborn University bearing (PU).}
The PU dataset \cite{Lessmeier2016} [Lessmeier 2016] is a well-known and frequently utilized dataset in the field of bearing analysis. It comprises recordings of high-frequency vibrations and motor currents from a total of 32 bearings, consisting of 26 faulty bearings and 6 healthy bearings. 

In addition to the vibration and current data, the dataset provides measurements of speed, load, torque, and temperature, offering a comprehensive set of information for analysis. The signals were collected under four different working conditions, each representing a distinct operating scenario. These working conditions are used for the LOGO cross-validation strategy in this study.

\paragraph{Naphthalene concentration (Naph).}
The Naph dataset \cite{Bastuck2015} consists of recordings from a gas sensor operating at different temperatures. The sensor signals were sampled at a rate of 4 Hz. The dataset focuses on the detection and analysis of naphthalene concentrations in the presence of ethanol as a background or interfering gas. Indeed, despite the dataset not being originally from the condition monitoring domain, it exhibits similarities with datasets used in condition monitoring, particularly in terms of signal shape and cross-influence variables. These similarities make it feasible to employ a meaningful LOGO validation approach for the study.

The dataset includes measurements of six different concentrations of naphthalene. These concentrations were repeated for three levels of ethanol, representing different interference scenarios. The dataset is designed to evaluate the sensor's performance in detecting and quantifying naphthalene concentrations accurately in the presence of varying ethanol levels. To evaluate the performance of algorithms and models, the dataset employs the LOGO cross-validation strategy using the ethanol concentrations as distinct groups.

\paragraph{Human activity recognition using smartphones (HAR).}
The HAR dataset \cite{Jorge2015} [Jorge 2015] consists of sensor data recorded by a smartphone from 30 individuals performing six different activities: "walking", "walking upstairs", "walking downstairs", "sitting", "standing", and "lying". The sensor data includes 3-axial linear acceleration and 3-axial angular velocity (gyro) captured at a constant rate of 50Hz.

Although the dataset does not originate from industrial applications, it holds significance in this study due to several reasons. The signal shapes, sensor types, and the target variable align with typical datasets used in the condition monitoring field. The objective of this dataset is to classify the activities performed by the participants based on the sensor signals. In this dataset only the z-axis signal of the linear accelerometer is used as the source of the data.

To evaluate the performance of the classification models, the participants are divided into four groups. The LOGO validation method is employed, where one group is left out for testing while the remaining three groups are used for training. This process is repeated for each group to ensure comprehensive evaluation.

\begin{table}
 \caption{Table showing the designed scenarios in the study.}
  \centering
  \begin{tabular}{ccccccc}
    \toprule
    \cmidrule(r){1-3}
    Dataset & K-Fold & LOGO & Num Observations & Signal Size & Num. Classes & Num. Domains\\
    \midrule
    CWRU & \checkmark & \checkmark & 1652 & 1024 & 4 & 4     \\
    HS (Acm) & \checkmark & \checkmark & 1449 & 6000 & 4 & 3     \\
    HS (Valve) & \checkmark & \checkmark & 1652 & 1024 & 4 & 4     \\
    EC & \checkmark & \checkmark & 11666 & 2000 & 5 & 4     \\
    OGW & \checkmark & \checkmark & 684 & 13108 & 2 & 4     \\
   
    PU & \checkmark & \checkmark & 1000 & 4096 & 2 & 4     \\
    Naph & \checkmark & \checkmark & 1569 & 160 & 6 & 3     \\
     HAR & \checkmark & \checkmark & 7352 & 128 & 4 & 4     \\
    \bottomrule
  \end{tabular}
  \label{tab:table}
\end{table}

Table \ref{tab:table} provides a summary of the dataset's features. The datasets share common characteristics, typically consisting of fewer than a few thousand observations, and the number of classes is relatively limited, with most having fewer than six classes. However, the data size can vary significantly, ranging from 100 to 10,000 data points.

\subsection{AutoML toolbox}
To evaluate the generated scenarios in this study, a MATLAB-based \cite{MATLAB2022} AutoML toolbox is employed. This toolbox utilizes an exhaustive search strategy to find the optimal combination of feature extraction (FE), feature selection (FS), and classification methods. The evaluation process includes testing various combinations of FE, FS, and classification techniques, as listed in Table \ref{tab:toolbox}.
\begin{table}
 \caption{Methods implemented in the AutoML toolbox. }
  \centering
  \begin{tabular}{ll}
    \toprule
    \cmidrule(r){1-2}
    Feature Extraction Methods\\
    \midrule
    ALA &  Adaptive linear approximation \cite{Olszewski2001}      \\
    BFC &  Best Fourier coefficient \cite{Mörchen2003}    \\
    BDW &	Best Daubechies wavelets \cite{Mörchen2003}  \\
    TFEx &	Statistical features in time and frequency domains \cite{Goodarzi2023} \\
    NoFE &	No feature extraction \\
    PCA &	Principal component analysis \cite{Wold1987} \\
    StatMom	& Statistical moments \cite{Schneider2018} \\
    \midrule
    Feature Selection Methods\\
    \midrule
    Pearson &	Pearson correlation coefficient \cite{freedman2007statistics}\\
    RELIEFF &	RELIEFF \cite{Kononenko1997} \\
    RFESVM &	Recursive feature elimination support vector machines \cite{lin2012support} \\
    Spearman &	Spearman correlation coefficient \cite{zar2005spearman}\\
    NoFS &	No feature selection\\
    \midrule
    Classification Methods \\
    \midrule
    LDAMahal &	Linear discriminant analysis with Mahalanobis distance classification \\
    SVM &	Support vector machine with a radial basis function kernel\\
    \bottomrule
  \end{tabular}
  \label{tab:toolbox}
\end{table}
 It is important to note that explicit hyperparameter tuning was not performed for each method in this study. Instead, the best number of features was determined using a brute force approach after the FS step. This approach helps identify the most relevant features for each method.

The AutoML toolbox used in this study has demonstrated impressive performance, as documented in multiple previous studies \cite{Schneider2018,Goodarzi2023, Schnur2022}. Its usage in this investigation is motivated not only by its performance but also by its focus on interpretability and explainability of the models \cite{Goodarzi2022}. In the context of predictive maintenance in industrial applications, interpretability is crucial for selecting a model that can be effectively deployed and understood by domain experts \cite{Hong2020}.

In this study, a selection of FE methods is employed to cover both the time and frequency domains. Specifically, methods such as ALA, PCA, and StatMom are utilized to extract features from the time domain using distinct approaches. BFC focuses on the frequency domain for FE, while BDW and TFEx extract features from both the time and frequency domains. Additionally, a NoFE approach is employed, which essentially does not perform any function on the data. This approach can be beneficial when dealing with short data instances or when the raw data itself already contains the necessary information for classification.

The comprehensive selection of FE methods allows for a thorough exploration of different feature representations, enabling the identification of the most informative features for each classification task.

\subsection{Deep learning methods}
In this research, we conduct a comparative analysis of four distinct neural network architectures: multi-layer perceptron (MLP) \cite{haykin1994neural}, convolutional neural network (CNN) \cite{schmidhuber2015deep}, Residual network (ResNet) \cite{He2016}, and WaveNet \cite{Oord2016}. We also evaluate their performance against the conventional AutoML toolbox. To achieve optimal network configurations, each architecture is subjected to hyper-parameter tuning. The specific parameters utilized for tuning each network are outlined in Table \ref{tab:HP}. Through this comprehensive evaluation, we aim to identify the most suitable neural network architecture for the given tasks.

Training neural networks can be computationally expensive, especially when combined with the challenge of hyperparameter tuning. To address this, early stopping and Bayesian optimization are employed to reduce the computational load. The maximum number of iterations for Bayesian optimization is set to 100, with a maximum time limit of 3600 seconds. In this study, we adopt Adam optimization as the optimization method for all networks, while employing the cross entropy loss function for the training process.

The training process is conducted using Nvidia RTX 5000 GPUs, with a maximum mini-batch size of 64 to accommodate hardware limitations.
\begin{figure} % picture
    \centering
    \includegraphics[width=\textwidth]{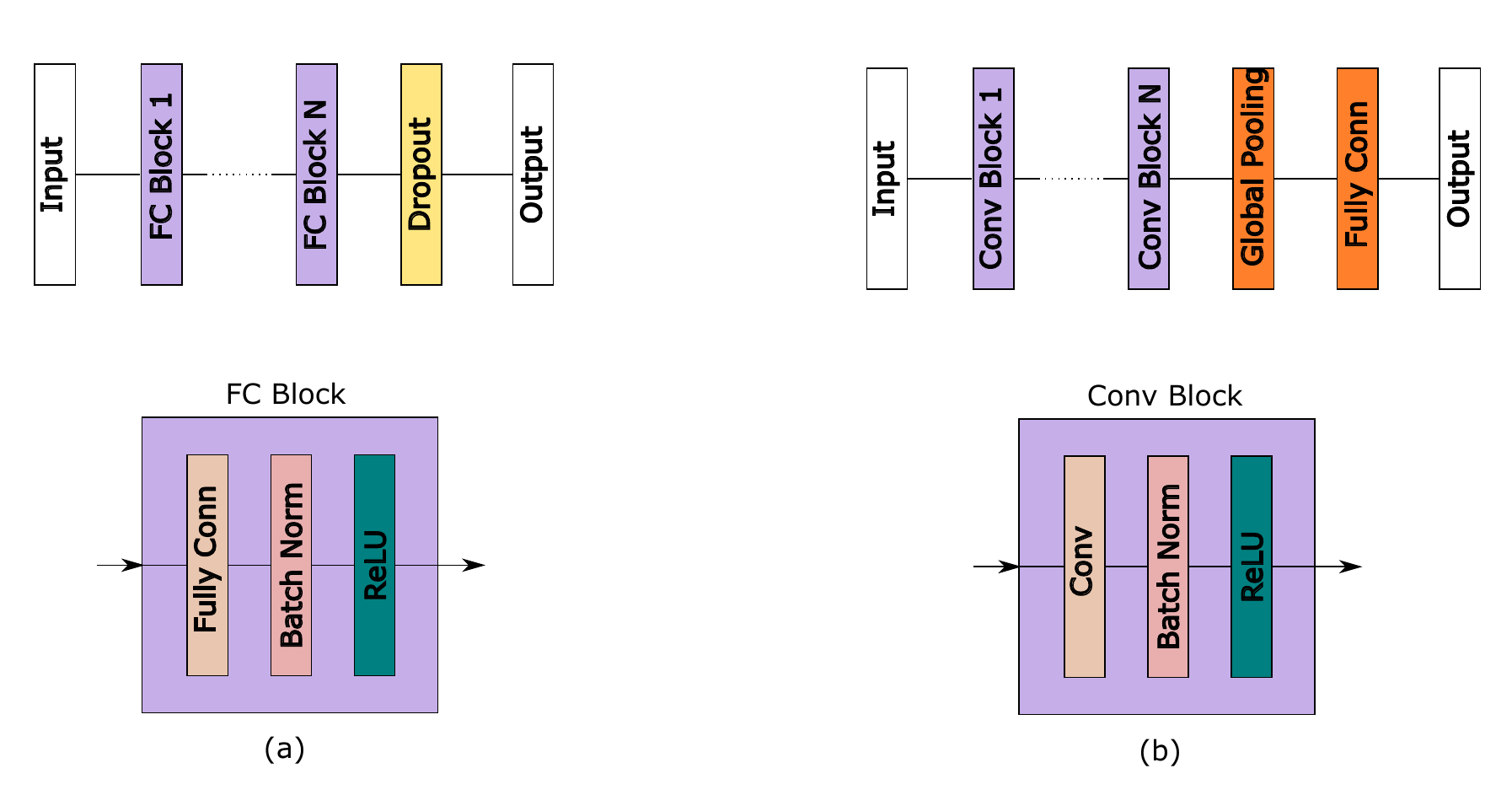}
      \caption{Structure of MLP (a) and CNN networks (b).}
  \label{fig:MLP}
\end{figure}
\paragraph{Multi-Layer Perceptron.}
The architecture of the MLP network comprises multiple blocks, with each block containing a sequence of layers including a fully connected layer, a batch normalization layer, and a ReLU activation layer. Figure \ref{fig:MLP}(a) provides a visual representation of the structure of the MLP network, showcasing the arrangement of its components within each block. The MLP architecture has been widely employed in various condition monitoring applications \cite{Avci2021,fawaz2019}, highlighting its relevance and applicability in the field. In the designed MLP network the number of neurons in each block follows the equation: $num\_neurons(i) = num\_neurons\_base/(2i-1)$, where $i$ represents the block index. The network architecture is defined by the number of blocks and $num\_neurons\_base$.

 To prevent overfitting, we include a dropout layer at the end of the network, which randomly drops out some neurons during training. Notably, the number of neurons in each block decreases as the depth of the network increases. This progressive shrinking of the number of neurons in deeper layers helps reduce the overall complexity of the model, mitigating the risk of overfitting and enhancing the generalization performance.
\paragraph{Convolutional Neural Network.}
The CNN architecture in this study consists of multiple convolutional blocks, each designed to extract features from the input data, Figure \ref{fig:MLP}(b). The first block has its own filter size and stride, which are independent of the subsequent blocks. However, all the remaining blocks share the same filter and stride size. The number of filters in each block is determined by multiplying the base number of filters by the block number.

After the convolutional layers, we employ global pooling, which aggregates the output of each feature map into a single value. This is followed by a fully connected layer with a variable number of neurons. The purpose of the fully connected layer is to perform classification based on the learned features.

The CNN network architecture is widely used in image classification tasks due to its ability to capture spatial relationships and extract relevant features from the input images through the convolutional layers. It is also widely used in time series classification \cite{Avci2021, fawaz2019,Kiranyaz2021}. The progressive increase in the number of filters in each block allows the network to learn more complex and abstract features as it goes deeper. The use of global pooling helps reduce the number of parameters in the fully connected layer, making the model more efficient and reducing the risk of overfitting. Additionally, the inclusion of dropout further mitigates overfitting by introducing randomness during training. The variable number of neurons in the fully connected layer provides flexibility in adjusting the complexity of the model to match the specific requirements of the task at hand.
\begin{equation}
     num\_filters=num\_filters\_base\times2^{cnn\_block-1}
\end{equation}
\begin{figure} % picture
    \centering
    \includegraphics[width=\textwidth]{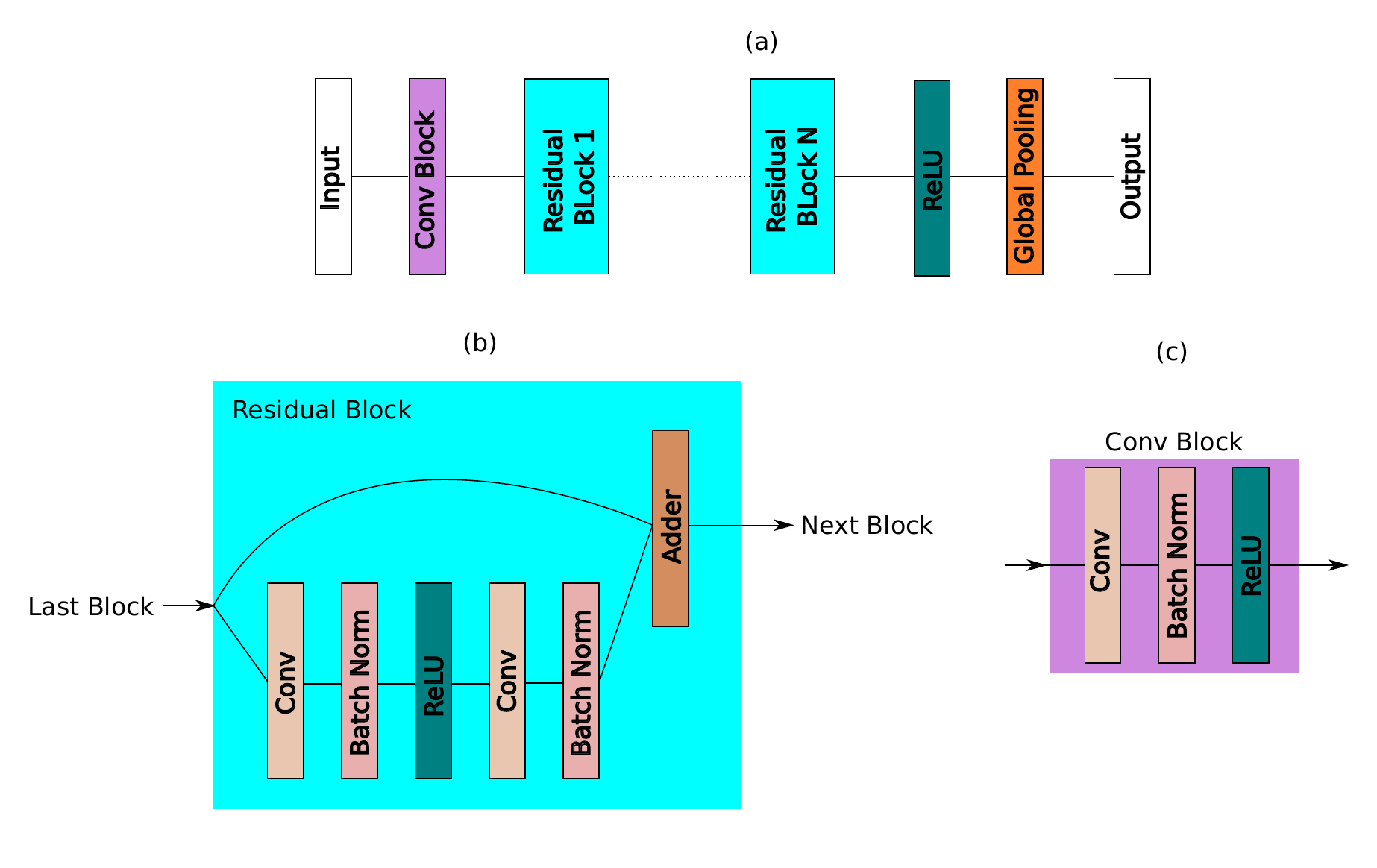}
      \caption{Structure of ResNet. (a) the main structure, (b) the residual block, and (c) the convolutional block.}
  \label{fig:ResNet}
\end{figure}
\paragraph{Residual Network.}
The ResNet architecture is designed to tackle the challenge of the vanishing gradient problem encountered in deep neural networks. To address this issue, ResNet introduces skip connections that enable the direct flow of information from one layer to another, bypassing intermediate layers. This approach helps to mitigate the degradation of performance that can occur as a network becomes deeper.

In the ResNet architecture, each block is composed of convolutional layers, batch normalization, and ReLU activation functions. The first block in each stage employs a larger filter size and stride size, which results in down-sampling of the input, Figure \ref{fig:ResNet}. This down-sampling is beneficial for capturing high-level abstract features and reducing the spatial dimensions of the input. Subsequent blocks within the same stage maintain the same filter size and stride size to preserve the feature map dimensions.

The skip connections in ResNet allow the network to learn residual mappings by modeling the difference between the desired output and the input. By incorporating these residual connections, the network can more effectively propagate gradients during training and improve the flow of information throughout the architecture. This enables the network to learn deeper and more expressive representations, leading to improved performance on challenging tasks \cite{fawaz2019}.

The number of filters in each block is determined by the equation:
\begin{equation}
    num\_filters=num\_filters\_base\times2^{stage-1}
\end{equation}
The total number of blocks is divided into stages, and each stage consists of a fixed number of blocks determined by the equation:
\begin{equation}
    num\_block\_per\_stage\ =\ floor(num\_Blocks\ /3)
\end{equation}

\paragraph{WaveNet.}
The WaveNet architecture \cite{Oord2016}, originally developed by Google for audio generation, has demonstrated its effectiveness in various domains including speech recognition, music synthesis, and vibration signal classification \cite{Zhuang2019}. WaveNet incorporates two distinctive features that contribute to its success.

The first feature is the use of dilated convolutions, which enables the model to capture long-term dependencies in the data. By applying convolutional filters with exponentially increasing dilation factors, WaveNet effectively expands the receptive field, allowing the network to capture patterns that span a larger context. This is particularly advantageous for tasks that require modeling temporal dependencies over extended sequences.

The second feature of WaveNet is the inclusion of skip connections, similar to the ResNet architecture. These skip connections facilitate the direct transmission of information between layers, mitigating the loss of information as the network grows deeper. By preserving and propagating relevant information, skip connections enhance the model's ability to learn complex representations and facilitate training of deep architectures.

\begin{figure} % picture
    \centering
    \includegraphics[width=\textwidth]{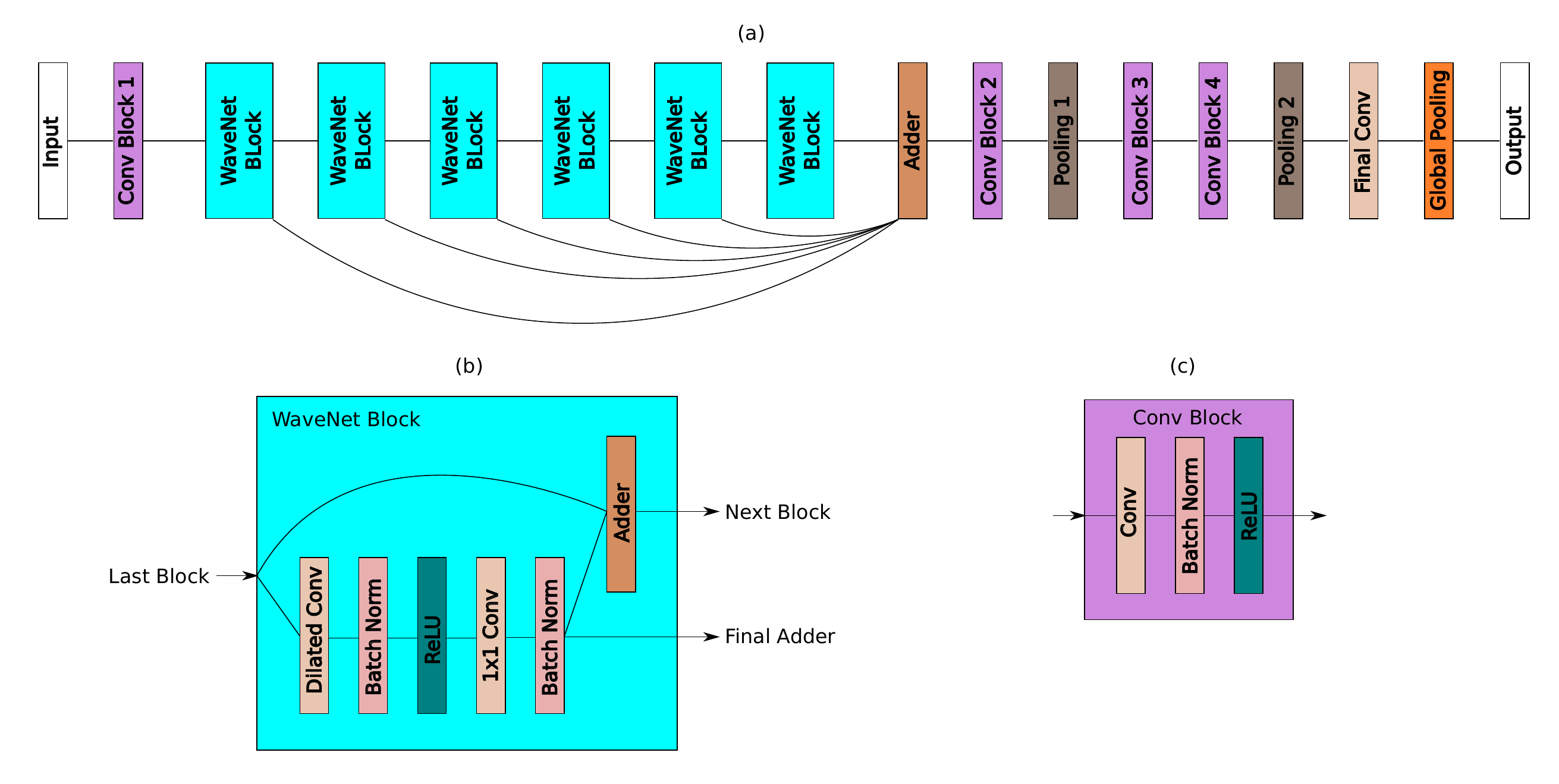}
      \caption{Structure of WaveNet. (a) the main structure, (b) the WaveNet block, and (c) the convolutional block.}
  \label{fig:WaveNet}
\end{figure}

In the current study, the WaveNet architecture is defined using key parameters such as the number of blocks, the filter length, and stride of the initial block, the filter size and number of filters in each subsequent block, and the pooling size of the final stage, Figure \ref{fig:WaveNet}. Additionally, the dilation factor follows the exponential growth pattern, as outlined in the original WaveNet paper. This configuration enables the WaveNet model to effectively capture long-term dependencies and leverage skip connections to enhance its representation learning capabilities.
\begin{table}
 \caption{Ranges of hyper-parameters for various network architectures.}
  \centering
  \begin{tabular}{lllll}
    \toprule
    \cmidrule(r){1-2}
    Network type &	Variable HPs &	Range &	Fixed HPs &	Value\\
    \midrule
    Common HPs &       \\
        &  Num. filters &	[8,128] & &     \\
        &  Num. FC/Conv/Res/Wave blocks &	[3,10] & &     \\
        &  Filter size &	[2,32] & &     \\
        &  Filter size, 1st layer &	[3,128] & &     \\
        &  First stride &	[1,50] & &     \\
    \midrule
    MLP&       \\
        &  Num. neurons &	[10,500] & Dropout & 40\%    \\
    \midrule
    CNN \\
     &  Num. neurons &	[100,500] \\
     &  Stride  &	[1,3] \\
     &  Global pooling &	[false,true] \\
    \midrule
    ResNet \\
     &  Stride 2nd stage &	[2,8] & Stride 1st stage &	1 \\
     &  Stride 3rd stage &	[2,8]  \\
     &  Global pooling &	[false,true] \\
        &  Num. neurons &	[10,500]\\
            \midrule
    WaveNet&       \\
        &  Pooling size &	[2,32] & Dropout & 40\%    \\
        \midrule
            Training HPs&       \\
       &Normalization &	[none,zscore] &	Mini batch size &	64   \\
        &Normalization dimension &	[element,all] &	Max epoch &	200  \\
        & &	 &	L2 regularization &	0.0001 \\
        & &	 &	Initial learning rate &	0.004 \\
        & &	 &	Learn rate drop factor &	0.8 \\
        & &	 &	Training time limit  &	60 sec \\
         & &	 &	Learn rate schedule &	piecewise \\
         \midrule
    Baysian Opt. \\
     &&&  Number of Trials &	100 \\
     &&&  Max Optimization Time  &	3000 sec \\
    \bottomrule
  \end{tabular}
  \label{tab:HP}
\end{table}

\subsection{Evaluation and model selection}
In this study, the evaluation metric used to assess the performance of the models is accuracy. Accuracy is a common metric employed in classification tasks and is defined as the ratio of correct predictions to the total number of observations. It provides an indication of how well the models are able to classify the data correctly.

Alternatively, accuracy (acc) can also be expressed as 1 minus the error rate (err). The error rate represents the expected value of the 0-1 loss across all observations, which is the loss incurred when a prediction does not match the true label. By subtracting the error rate from 1, accuracy provides a measure of the proportion of correct predictions.

In this study, the evaluation metric used to assess the performance of the models is accuracy. This metric provides a measure of the overall correctness of the models' predictions and serves as a clear indicator of their classification performance. By considering the accuracy, the study aims to analyze and compare the models' abilities to correctly classify the given data.
\begin{equation}
    err=\frac{1}{n}\sum_{i=1}^{n}\left[y_i\neq f\left(x_i\right)\right]
\end{equation}
\begin{equation}
    acc=1-err
\end{equation}

\subsection{Occlusion map}
Occlusion maps \cite{zeiler2014visualizing} are a post-hoc method that provides valuable insights into the decision-making process of neural networks, which are often considered black boxes due to their lack of interpretability. In industrial applications and other scenarios where explanations for decisions are crucial, occlusion maps serve as a reliable tool. This method involves applying the occlusion technique to a trained network, treating each observation individually. 

To generate an occlusion map, the input data is perturbed using a sliding mask, which covers different regions of the input. By measuring the network's sensitivity to these perturbations, an attribution map is obtained, indicating the importance of various regions in the input for making decisions. This map offers insights into which parts of the input are influential in the network's decision-making process.

To effectively perform occlusion mapping, three key parameters need to be carefully defined. The first parameter is the mask size, which determines the size of the sliding window used to occlude different regions of the input. The second parameter is the stride size, which determines the step size for sliding the mask over the input. Finally, the mask value is the value that replaces the original input value during the perturbation process.

Overall, occlusion maps provide an intuitive and easily interpretable approach to understanding the decisions made by neural networks. They offer valuable insights into the model's behavior and enable better explanations for its decisions, making them particularly useful in scenarios where interpretability is important, such as industrial applications.
\section{Results}
\begin{figure} % picture
    \centering
    \includegraphics[width=\textwidth]{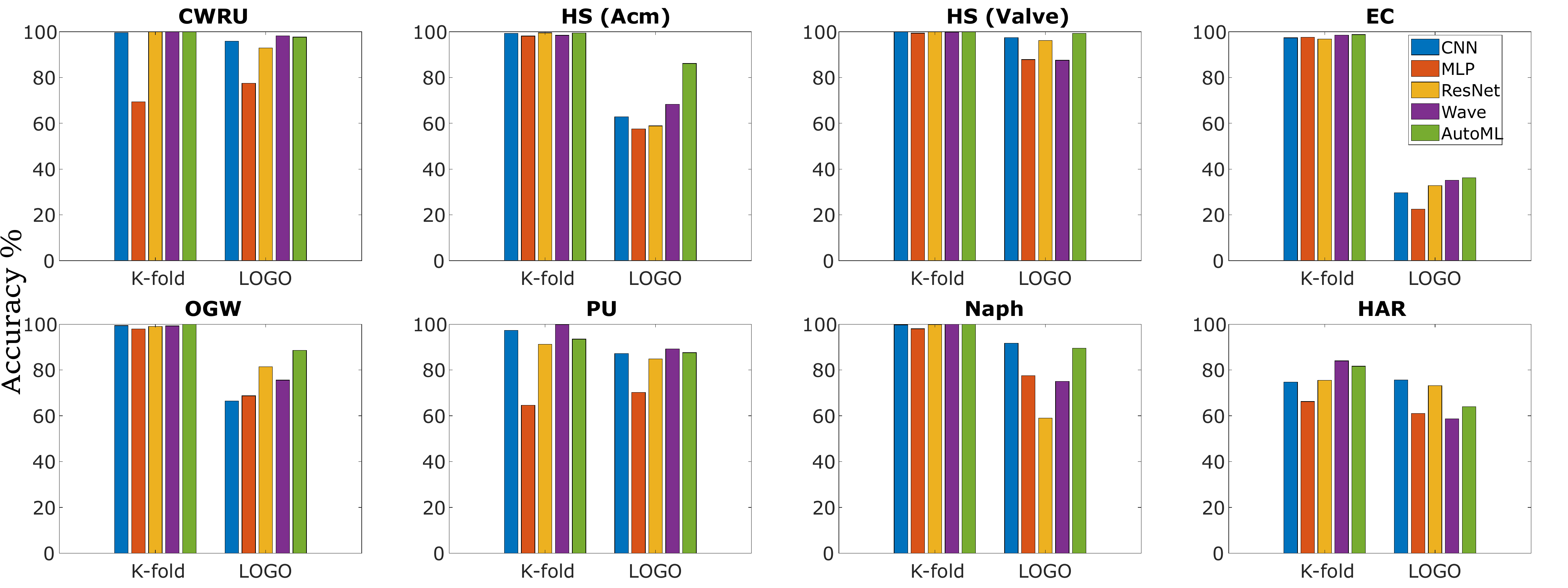}
      \caption{Comparing the accuracy of various networks and the AutoML toolbox in both K-fold and LOGO scenarios.}
  \label{fig:Methods}
  \end{figure}
  
The evaluation using the AutoML toolbox involved testing 70 (7 FE methods x 5 FS methods x 2 classification algorithms) combinations of methods for the desired tasks. The selection of the best method was based on the accuracy of the models, which were assessed through cross-validation.

In Figure \ref{fig:Methods}, a bar graph displays the accuracies of the evaluated models, divided into two groups: LOGO and K-fold validations. For the majority of use-cases (6 out of 8), all models demonstrate almost perfect accuracy with K-fold validation, posing a challenge in distinguishing their performance differences. Conversely, LOGO validation reveals a different scenario, where only one use-case achieves near-perfect accuracy, which is more interesting. LOGO validation highlights the issue of domain shift, emphasizing the lack of a clear winner among the tested models. Notably, in all use-cases except HAR, the toolbox exhibits the highest accuracy or remains as the runner-up with only marginal differences. While no definitive winner emerges among the tested DNN architectures, it is worth noting that in use-cases involving vibration signals, MLP consistently ranks at the bottom.

The MLP network had the lowest overall mean accuracy among all the tested networks, indicating poorer performance compared to the others. On the other hand, the CNN network exhibited the highest overall accuracy among the tested networks. Since ResNet and WaveNet are also CNNs, the flexible parametric design of CNNs may have contributed to their superior performance, enabling them to handle both short and long signals effectively. However, it is worth noting that other studies have demonstrated that ResNet can outperform simpler CNN architectures when the number of layers increases. It is worth mentioning that WaveNet outperformed the other networks for the two datasets that contained vibration data. This observation is consistent with the capabilities of the WaveNet architecture in handling long input signals and utilizing receptive fields with varying sizes, which are advantageous for capturing relevant information in vibration data.
\begin{figure} % picture
    \centering
    \includegraphics[width=\textwidth]{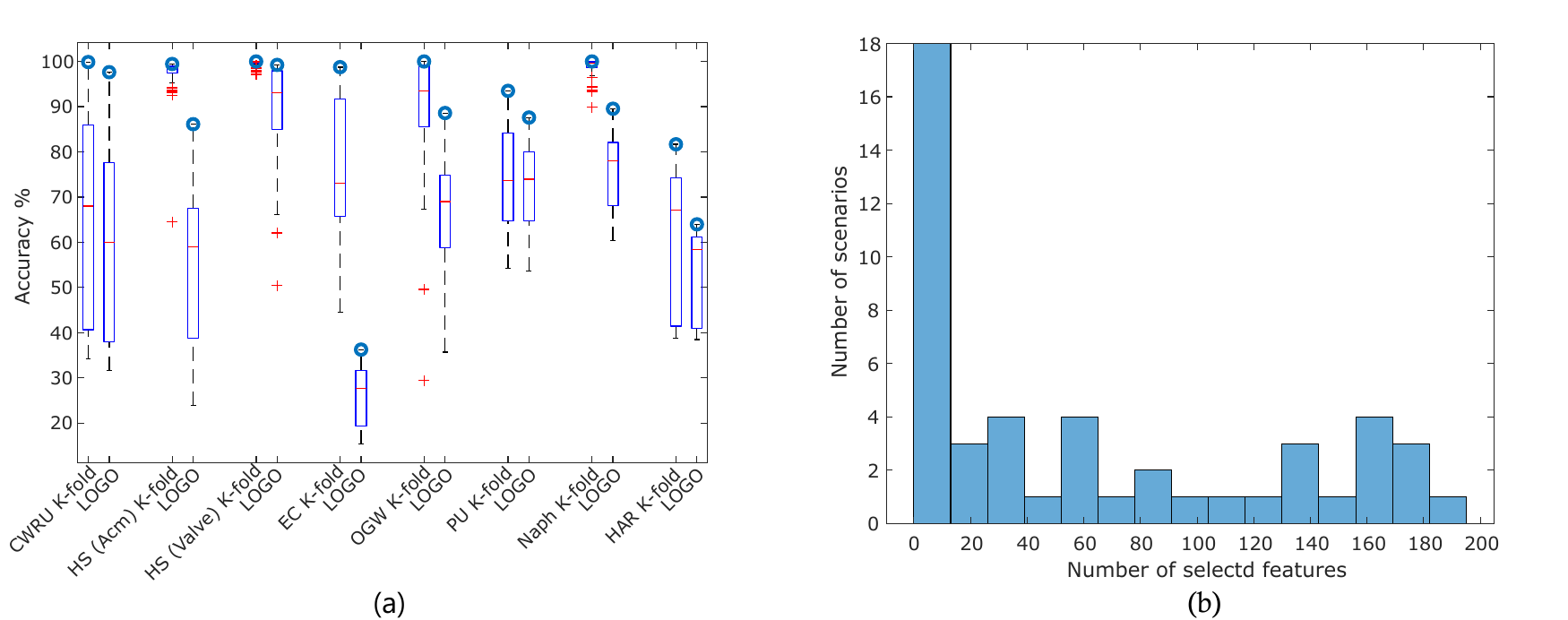}
      \caption{ (a) The caption shows a visual comparison of box plots, depicting the errors of conventional models from the AutoML toolbox in both LOGO and K-fold validation scenarios. For each task, the model with the highest cross-validation accuracy is chosen.
(b) Histogram depicting the distribution of the number of selected features for the top three models.}
  \label{fig:box plot}
  \end{figure}

Figure \ref{fig:Methods}(a) just shows the accuracy of the best model however AutoML toolbox explores variety of models on each task. To have a close look at the methods from AutoML, box plots of cross-validation errors from the toolbox models are shown in Figure \ref{fig:box plot}. As in line with the previous results, it can be observed that the LOGO validation scenarios consistently resulted in higher errors compared to the K-fold scenarios. Additionally, the errors in the LOGO validation scenarios exhibited higher variation, indicating the importance of careful model selection in these scenarios and real-world applications. Notably, for the HS dataset, the tested models achieved low error rates. However, there was a significant difference of approximately 80\% between the best and worst error rates, highlighting the significance of selecting the appropriate model for this dataset.

In the subsequent section, we conduct a comprehensive comparison between the neural network models and the AutoML approach, beginning with hyperparameter optimization. Hyperparameter optimization is a crucial and resource-intensive phase in the model design process, requiring substantial time and computational resources. As detailed in section \ref{methods} of our paper, we meticulously fine-tuned the hyperparameters for the neural networks across an extensive range of parameters. However, we did not explicitly perform fine-tuning for the hyperparameters of the AutoML toolbox methods. The only parameter adjusted during the toolbox's cross-validation was the optimal number of features, determined after employing the feature selection method.

NNs and conventional methods can be compared from another perspective. Some conventional methods from the AutoML toolbox may be affected by the number of observations and classes. For instance, the training time of one vs one SVM is quadratic proportional to the number of classes because each binary classifier is trained on a pair of classes, requiring a number of binary classifiers equal to the number of pairs of classes. This number grows quadratically with the number of classes, with N(N-1)/2 pairs of classes for N classes. Conversely, training NNs using back propagation and specialized frameworks designed to utilize high-end GPUs can handle very large numbers of observations faster and more efficiently.

In many condition monitoring applications, it is often necessary to extract only a few features from the input signal to perform various tasks. Figure \ref{fig:box plot}(b) provides a histogram of the selected features of the top-3 methods of the toolbox for each scenario, which reveals that in many scenarios, the best results are achieved using fewer than 20 features of the signal. The limited number of classes in each task is one reason for this phenomenon. In many supervised learning tasks related to condition monitoring, the number of classes can be reduced to only two classes: healthy and faulty cases. This is in contrast to AudioSet \cite{Gemmeke2017}, which has more than 500 classes, or the well-known ImageNet \cite{DenDon09Imagenet}, which has 1000 classes. With a limited number of observations, target classes and necessary features, usually low-capacity models are sufficient to perform the tasks.
 \begin{figure} % picture
    \centering
    \includegraphics[width=\textwidth]{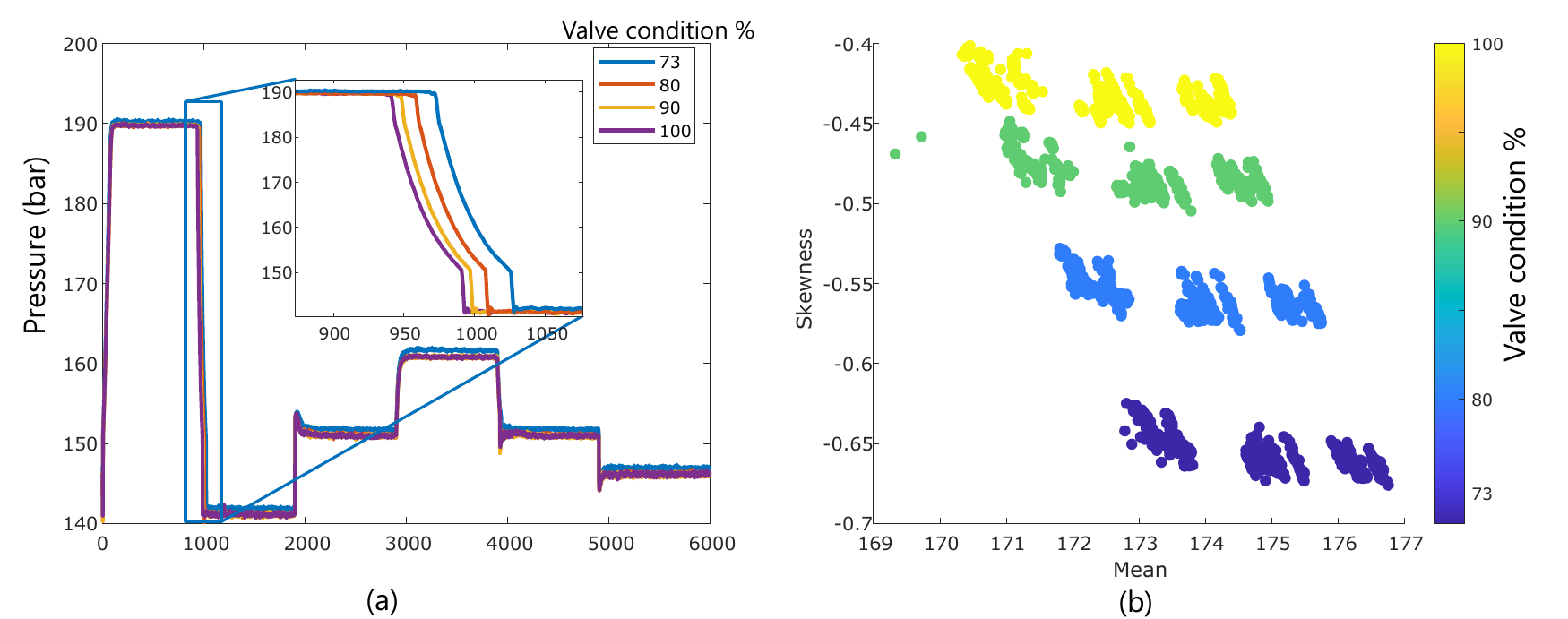}
      \caption{(a) Four representative observations from the HS-valve dataset. (b) Key features identified by the best model as the most important.}
  \label{fig:features}
  \end{figure}
The conventional methods available in the AutoML toolbox offer a significant advantage in terms of interpretability. These methods are structured as separate blocks, each serving specific functions, which facilitates the identification of crucial features. This interpretability is particularly enhanced when a limited number of features are used, and linear methods are employed. For instance, in the HS (Valve) scenario with LOGO validation, the most accurate model utilized statistical moments as the feature extractor, Pearson correlation as the feature selector, and SVM as the classifier. Notably, only the first drop in the signal (depicted in Figure \ref{fig:features}) was attributed to the valve's transaction, while the subsequent steps were related to the designed process. During the classification step, only four features were employed from the extracted features. By retracing these features to the raw signal, it becomes possible to pinpoint the significant regions in the signal, specifically the switching region of the valve. A clear visualization of the efficacy of two selected features is illustrated in Figure \ref{fig:features}(b), revealing three distinct groups corresponding to each class label, representing the cooler performance utilized in the LOGO validation scenario. 

In contrast, explaining the decisions made by neural networks is a more challenging task, and various methods have been proposed to interpret their predictions. One approach to identify important features is through the use of occlusion maps, which provide a visualization of the crucial regions in the input data. For comparison purposes, we selected a CNN model as it demonstrated the best results in the HS (Valve) scenario. However, interpreting the results from occlusion maps also involves selecting appropriate parameters such as the mask size, mask stride, and replacing value. Since there are no true labels for the attributions, we used prior knowledge about the use-case from the classical methods and performed a grid search to find suitable parameters.

Utilizing our prior knowledge of the use-case, we imposed certain constraints on the parameters. For instance, to prevent covering two transactions of the signal and to ensure accurate identification of important features, we set the mask size not to exceed 1000. Additionally, we aimed for a mask size small enough to facilitate feature identification effectively. To explore the best parameter values, we conducted a grid search over the mask size and stride, ranging from 5 to 1000.

\begin{figure} % picture
    \centering
    \includegraphics[width=\textwidth]{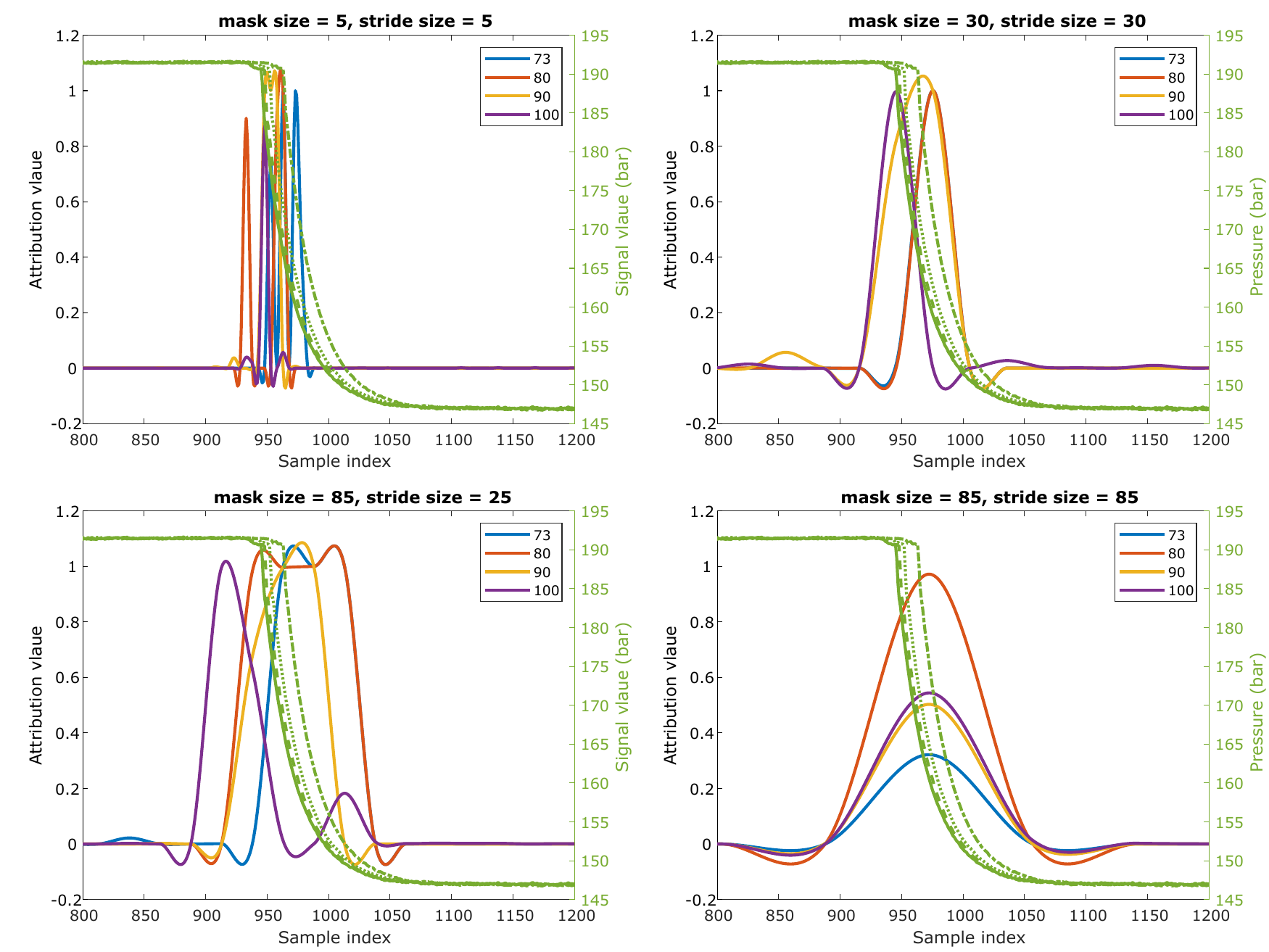}
      \caption{Four different combinations of mask size and stride settings for generating occlusion maps.}
  \label{fig:maps}
  \end{figure}

In Figure \ref{fig:maps}, we showcase the attribution maps of the observations with 100\% cooling performance for three sample values of mask sizes and four stride sizes. While the peak and its position in the attribution maps may differ with the selection of parameters, it is consistent that the network consistently emphasizes a similar region in the input data as the most important feature (further details on how this is demonstrated will be provided in the paper). However, precisely identifying the specific features within that highlighted region remains a challenge and warrants further investigation.

  \section{Conclusion}
  In conclusion, this research offers a comprehensive study that compares AutoML employing both conventional ML methods and DNN models for condition monitoring tasks.

The results of the experiments reveal several noteworthy insights. Firstly, the K-fold validation consistently demonstrates high accuracy for all models, except for MLP, which faces challenges in effectively capturing time-varying data, such as vibrations. On the other hand, the LOGO validation scenario for the same tasks does not lead to a definitive winner among the tested models. Nevertheless, it is worth mentioning that the conventional methods from the AutoML toolbox consistently achieve the highest accuracy or closely compete as the runner-up, with only minor differences between them.

Moreover, the comparison between conventional methods and neural networks provided further insights. Conventional methods from the AutoML toolbox, despite their interpretability and ease of identifying important features, were shown to be affected by the number of observations and classes. In contrast, DNNs trained using backpropagation and specialized frameworks designed for high-end GPUs demonstrated the capability to handle large numbers of observations more efficiently.

The experiments also highlighted the significance of feature selection in condition monitoring tasks. It was observed that, in many scenarios, the best results were achieved using fewer than 20 features extracted from the input signal. This phenomenon can be attributed to the limited number of classes typically encountered in condition monitoring, where the focus is often on distinguishing between healthy and faulty cases. Consequently, low-capacity models were found to be sufficient for these tasks.

In summary, the conducted experiments have underscored the significance of selecting the appropriate validation scenario to accurately evaluate ML models in realistic conditions. Notably, employing Random K-fold validation has proven to conceal the issue of domain shift, a prevalent challenge in condition monitoring tasks. Additionally, the study has highlighted the importance of feature selection, which has proven instrumental in achieving outstanding results with low-capacity models. It is noteworthy that utilizing suitable conventional ML methods has often resulted in superior or comparable outcomes to those attained by DNN models, without the added burden of interpreting complexities. The contributions of this study offer potential for pushing the field forward and improving the effectiveness of ML techniques, providing valuable direction in selecting the most appropriate models for practical industrial applications.

\bibliographystyle{IEEEtran}
\bibliography{ref}

\begin{thebibliography}{10}
\providecommand{\url}[1]{#1}
\csname url@samestyle\endcsname
\providecommand{\newblock}{\relax}
\providecommand{\bibinfo}[2]{#2}
\providecommand{\BIBentrySTDinterwordspacing}{\spaceskip=0pt\relax}
\providecommand{\BIBentryALTinterwordstretchfactor}{4}
\providecommand{\BIBentryALTinterwordspacing}{\spaceskip=\fontdimen2\font plus
\BIBentryALTinterwordstretchfactor\fontdimen3\font minus
  \fontdimen4\font\relax}
\providecommand{\BIBforeignlanguage}[2]{{%
\expandafter\ifx\csname l@#1\endcsname\relax
\typeout{** WARNING: IEEEtran.bst: No hyphenation pattern has been}%
\typeout{** loaded for the language `#1'. Using the pattern for}%
\typeout{** the default language instead.}%
\else
\language=\csname l@#1\endcsname
\fi
#2}}
\providecommand{\BIBdecl}{\relax}
\BIBdecl

\bibitem{gertsbakh2000reliability}
I.~Gertsbakh and I.~B. Gertsbakh, \emph{Reliability theory: with applications
  to preventive maintenance}.\hskip 1em plus 0.5em minus 0.4em\relax Springer
  Science \& Business Media, 2000.

\bibitem{mobley2002introduction}
R.~K. Mobley, \emph{An introduction to predictive maintenance}.\hskip 1em plus
  0.5em minus 0.4em\relax Elsevier, 2002.

\bibitem{Olah2017}
\BIBentryALTinterwordspacing
C.~Olah, A.~Mordvintsev, and L.~Schubert, ``Feature visualization,''
  \emph{Distill}, vol.~2, 11 2017. [Online]. Available:
  \url{distill.pub/2017/feature-visualization}
\BIBentrySTDinterwordspacing

\bibitem{Schneider2018}
\BIBentryALTinterwordspacing
T.~Schneider, N.~Helwig, and A.~Schütze, ``Industrial condition monitoring
  with smart sensors using automated feature extraction and selection,''
  \emph{Measurement Science and Technology}, vol.~29, p. 94002, 8 2018.
  [Online]. Available: \url{https://doi.org/10.1088/1361-6501/aad1d4}
\BIBentrySTDinterwordspacing

\bibitem{Avci2021}
O.~Avci, O.~Abdeljaber, S.~Kiranyaz, M.~Hussein, M.~Gabbouj, and D.~J. Inman,
  ``A review of vibration-based damage detection in civil structures: From
  traditional methods to machine learning and deep learning applications,''
  \emph{Mechanical Systems and Signal Processing}, vol. 147, p. 107077, 1 2021.

\bibitem{zhao2019}
R.~Zhao, R.~Yan, Z.~Chen, K.~Mao, P.~Wang, and R.~X. Gao, ``Deep learning and
  its applications to machine health monitoring,'' \emph{Mechanical Systems and
  Signal Processing}, vol. 115, pp. 213--237, 1 2019.

\bibitem{Selvaraju2020}
R.~R. Selvaraju, M.~Cogswell, A.~Das, R.~Vedantam, D.~Parikh, and D.~Batra,
  ``Grad-cam: Visual explanations from deep networks via gradient-based
  localization,'' \emph{International Journal of Computer Vision}, vol. 128,
  pp. 336--359, 2 2020.

\bibitem{Truong2019}
A.~Truong, A.~Walters, J.~Goodsitt, K.~Hines, C.~B. Bruss, and R.~Farivar,
  ``Towards automated machine learning: Evaluation and comparison of automl
  approaches and tools.''\hskip 1em plus 0.5em minus 0.4em\relax IEEE ICTAI, 11
  2019, pp. 1471--1479.

\bibitem{He2021}
X.~He, K.~Zhao, and X.~Chu, ``Automl: A survey of the state-of-the-art,''
  \emph{Knowledge-Based Systems}, vol. 212, p. 106622, 1 2021.

\bibitem{Esakimuthu2019}
S.~E. Pandarakone, Y.~Mizuno, and H.~Nakamura, ``A comparative study between
  machine learning algorithm and artificial intelligence neural network in
  detecting minor bearing fault of induction motors,'' \emph{Energies},
  vol.~12, p. 2105, 1 2019.

\bibitem{Buckley2022}
\BIBentryALTinterwordspacing
T.~Buckley, B.~Ghosh, and V.~Pakrashi, ``A feature extraction \& selection
  benchmark for structural health monitoring,'' \emph{Structural Health
  Monitoring}, vol.~22, pp. 2082--2127, 2022. [Online]. Available:
  \url{https://doi.org/10.1177/14759217221111141}
\BIBentrySTDinterwordspacing

\bibitem{fawaz2019}
H.~I. Fawaz, G.~Forestier, J.~Weber, L.~Idoumghar, and P.-A. Muller, ``Deep
  learning for time series classification: a review,'' \emph{Data Mining and
  Knowledge Discovery}, vol.~33, pp. 917--963, 2019.

\bibitem{Goodarzi2022}
\BIBentryALTinterwordspacing
P.~Goodarzi, A.~Schütze, and T.~Schneider, ``Comparison of different ml
  methods concerning prediction quality, domain adaptation and robustness,''
  \emph{Technisches Messen}, vol.~89, pp. 224--239, 2022. [Online]. Available:
  \url{https://doi.org/10.1515/teme-2021-0129}
\BIBentrySTDinterwordspacing

\bibitem{CWRU}
\BIBentryALTinterwordspacing
``Case western reserve university bearing data set.'' [Online]. Available:
  \url{https://engineering.case.edu/bearingdatacenter}
\BIBentrySTDinterwordspacing

\bibitem{Schneider2018HS}
T.~Schneider, S.~Klein, and M.~Bastuck, ``Condition monitoring of hydraulic
  systems data set at zema,'' \emph{Zenodo}, 4 2018.

\bibitem{Klein2018}
\BIBentryALTinterwordspacing
S.~Klein, ``Sensor data set, electromechanical cylinder at zema testbed
  (2khz),'' \emph{Zenodo}, 8 2018. [Online]. Available:
  \url{https://zenodo.org/record/3929384}
\BIBentrySTDinterwordspacing

\bibitem{Moll2019}
\BIBentryALTinterwordspacing
J.~Moll, C.~Kexel, S.~Pötzsch, M.~Rennoch, and A.~S. Herrmann, ``Temperature
  affected guided wave propagation in a composite plate complementing the open
  guided waves platform,'' \emph{Scientific Data}, vol.~6, p. 191, 2019.
  [Online]. Available: \url{https://doi.org/10.1038/s41597-019-0208-1}
\BIBentrySTDinterwordspacing

\bibitem{Lessmeier2016}
C.~Lessmeier, J.~K. Kimotho, D.~Zimmer, and W.~Sextro, ``Condition monitoring
  of bearing damage in electromechanical drive systems by using motor current
  signals of electric motors: A benchmark data set for data-driven
  classification,'' 2016.

\bibitem{Bastuck2015}
M.~Bastuck, M.~Leidinger, T.~Sauerwald, and A.~Schütze, ``Improved
  quantification of naphthalene using non-linear partial least squares
  regression,'' \emph{ISOEN 2015}, 7 2015.

\bibitem{Jorge2015}
J.-L. Reyes-Ortiz, L.~Oneto, A.~SamÃ, X.~Parra, and D.~Anguita,
  ``Transition-aware human activity recognition using smartphones,''
  \emph{Neurocomputing. Springer}, 2015.

\bibitem{MATLAB2022}
\BIBentryALTinterwordspacing
T.~M. Inc., ``Matlab version: 9.13.0 (r2022b),'' Natick, Massachusetts, United
  States, 2022. [Online]. Available: \url{https://www.mathworks.com}
\BIBentrySTDinterwordspacing

\bibitem{Olszewski2001}
R.~T. Olszewski, R.~Maxion, and D.~Siewiorek, ``Generalized feature extraction
  for structural pattern recognition in time-series data,'' \emph{PhD thesis,
  School of Computer Science, Carnegie Mellon University}, 2001, aAI3040489.

\bibitem{Mörchen2003}
F.~Mörchen, ``Time series feature extraction for data mining using dwt and
  dft,'' \emph{Tech. Rep.}, vol.~33, 12 2003.

\bibitem{Goodarzi2023}
P.~Goodarzi, S.~Klein, A.~Schütze, and T.~Schneider, ``Comparing different
  feature extraction methods in condition monitoring applications,'' 5 2023.

\bibitem{Wold1987}
\BIBentryALTinterwordspacing
S.~Wold, K.~Esbensen, and P.~Geladi, ``Principal component analysis,''
  \emph{Chemometrics and Intelligent Laboratory Systems}, vol.~2, pp. 37--52,
  1987, proceedings of the Multivariate Statistical Workshop for Geologists and
  Geochemists. [Online]. Available:
  \url{https://www.sciencedirect.com/science/article/pii/0169743987800849}
\BIBentrySTDinterwordspacing

\bibitem{freedman2007statistics}
D.~Freedman, R.~Pisani, and R.~Purves, ``Statistics (international student
  edition),'' \emph{Pisani, R. Purves, 4th edn. WW Norton \& Company, New
  York}, 2007.

\bibitem{Kononenko1997}
I.~Kononenko, E.~Šimec, and M.~Robnik-Šikonja, ``Overcoming the myopia of
  inductive learning algorithms with relieff,'' \emph{Applied Intelligence},
  vol.~7, pp. 39--55, 1997.

\bibitem{lin2012support}
X.~Lin, F.~Yang, L.~Zhou, P.~Yin, H.~Kong, W.~Xing, X.~Lu, L.~Jia, Q.~Wang, and
  G.~Xu, ``A support vector machine-recursive feature elimination feature
  selection method based on artificial contrast variables and mutual
  information,'' \emph{Journal of chromatography B}, vol. 910, pp. 149--155,
  2012.

\bibitem{zar2005spearman}
J.~H. Zar, ``Spearman rank correlation,'' \emph{Encyclopedia of Biostatistics},
  vol.~7, 2005.

\bibitem{Schnur2022}
C.~Schnur, P.~Goodarzi, Y.~Lugovtsova, J.~Bulling, J.~Prager, K.~Tschöke,
  J.~Moll, A.~Schütze, and T.~Schneider, ``Towards interpretable machine
  learning for automated damage detection based on ultrasonic guided waves,''
  \emph{Sensors}, vol.~22, p. 406, 1 2022.

\bibitem{Hong2020}
S.~R. Hong, J.~Hullman, and E.~Bertini, ``Human factors in model
  interpretability: Industry practices, challenges, and needs,''
  \emph{Proceedings of the ACM on Human-Computer Interaction}, vol.~4, pp.
  1--26, 5 2020.

\bibitem{haykin1994neural}
S.~Haykin, \emph{Neural networks: a comprehensive foundation}.\hskip 1em plus
  0.5em minus 0.4em\relax Prentice Hall PTR, 1994.

\bibitem{schmidhuber2015deep}
J.~Schmidhuber, ``Deep learning in neural networks: An overview,'' \emph{Neural
  networks}, vol.~61, pp. 85--117, 2015.

\bibitem{He2016}
K.~He, X.~Zhang, S.~Ren, and J.~Sun, ``Deep residual learning for image
  recognition,'' \emph{Proceedings of the IEEE Conference on Computer Vision
  and Pattern Recognition (CVPR)}, 6 2016.

\bibitem{Oord2016}
A.~van~den Oord, S.~Dieleman, H.~Zen, K.~Simonyan, O.~Vinyals, A.~Graves,
  N.~Kalchbrenner, A.~Senior, and K.~Kavukcuoglu, ``Wavenet: A generative model
  for raw audio,'' \emph{arXiv preprint arXiv:1609.03499}, 9 2016.

\bibitem{Kiranyaz2021}
S.~Kiranyaz, O.~Avci, O.~Abdeljaber, T.~Ince, M.~Gabbouj, and D.~J. Inman, ``1d
  convolutional neural networks and applications: A survey,'' \emph{Mechanical
  Systems and Signal Processing}, vol. 151, p. 107398, 4 2021.

\bibitem{Zhuang2019}
Z.~ZiLong, H.~Lv, J.~Xu, H.~Zizhao, and W.~Qin, ``A deep learning method for
  bearing fault diagnosis through stacked residual dilated convolutions,''
  \emph{Applied Sciences}, vol.~9, p. 1823, 05 2019.

\bibitem{zeiler2014visualizing}
M.~D. Zeiler and R.~Fergus, ``Visualizing and understanding convolutional
  networks,'' in \emph{Computer Vision--ECCV 2014: 13th European Conference,
  Zurich, Switzerland, September 6-12, 2014, Proceedings, Part I 13}.\hskip 1em
  plus 0.5em minus 0.4em\relax Springer, 2014, pp. 818--833.

\bibitem{Gemmeke2017}
J.~F. Gemmeke, D.~P.~W. Ellis, D.~Freedman, A.~Jansen, W.~Lawrence, R.~C.
  Moore, M.~Plakal, and M.~Ritter, ``Audio set: An ontology and human-labeled
  dataset for audio events,'' in \emph{2017 IEEE International Conference on
  Acoustics, Speech and Signal Processing (ICASSP)}, 2017, pp. 776--780.

\bibitem{DenDon09Imagenet}
\BIBentryALTinterwordspacing
J.~Deng, W.~Dong, R.~Socher, L.-J. Li, K.~Li, and L.~Fei-Fei, ``Imagenet: A
  large-scale hierarchical image database,'' in \emph{Computer Vision and
  Pattern Recognition, 2009. CVPR 2009. IEEE Conference on}.\hskip 1em plus
  0.5em minus 0.4em\relax IEEE, 2009, pp. 248--255. [Online]. Available:
  \url{https://ieeexplore.ieee.org/abstract/document/5206848/}
\BIBentrySTDinterwordspacing

\end{thebibliography}

\end{document}